\documentclass[conference]{IEEEtran}
\IEEEoverridecommandlockouts
\usepackage{cite}
\usepackage{amsmath,amssymb,amsfonts}
\usepackage{algorithmic} 
\usepackage{graphicx}
\usepackage{textcomp}
\usepackage{xcolor}
\usepackage{tikz}
\usepackage{lipsum}
\usetikzlibrary{arrows.meta, positioning,backgrounds}
\usepackage{comment}
\usepackage{adjustbox} 
\usepackage{makecell}  

\def\BibTeX{{\rm B\kern-.05em{\sc i\kern-.025em b}\kern-.08em
    T\kern-.1667em\lower.7ex\hbox{E}\kern-.125emX}}

\begin{document}

\title{TikGuard: A Deep Learning Transformer-Based Solution for Detecting Unsuitable TikTok Content for Kids}

\author{
\IEEEauthorblockN{Mazen Balat}
\IEEEauthorblockA{CS and IT\\
E-JUST\\
Alexandria, Egypt\\
mazen.balat@ejust.edu.eg}
\and
\IEEEauthorblockN{Mahmoud Essam Gabr}
\IEEEauthorblockA{Computers and Data Science\\
Alexandria University\\
Alexandria, Egypt\\
cds.mahmoudessam60231@alexu.edu.eg}
\and
\IEEEauthorblockN{Hend Bakr}
\IEEEauthorblockA{CS and IT\\
E-JUST\\
Alexandria, Egypt\\
hend.adel@ejust.edu.eg}
\and
\IEEEauthorblockN{Ahmed B. Zaky}
\IEEEauthorblockA{CS and IT\\
E-JUST\\
Alexandria, Egypt\\
ahmed.zaky@ejust.edu.eg}
}

\maketitle
\begin{abstract}
The rise of short-form videos on platforms like TikTok has brought new challenges in safeguarding young viewers from inappropriate content. Traditional moderation methods often fall short in handling the vast and rapidly changing landscape of user-generated videos, increasing the risk of children encountering harmful material. This paper introduces TikGuard, a transformer-based deep learning approach aimed at detecting and flagging content unsuitable for children on TikTok. By using a specially curated dataset, TikHarm, and leveraging advanced video classification techniques, TikGuard achieves an accuracy of 86.7\%, showing a notable improvement over existing methods in similar contexts. While direct comparisons are limited by the uniqueness of the TikHarm dataset, TikGuard’s performance highlights its potential in enhancing content moderation, contributing to a safer online experience for minors. This study underscores the effectiveness of transformer models in video classification and sets a foundation for future research in this area.

\end{abstract}


\section{Introduction}

In today's digital age, where short-form videos dominate social media platforms, safeguarding young viewers has become a critical challenge. TikTok, one of the most popular platforms among children and teenagers, presents a unique set of challenges due to the sheer volume and dynamic nature of user-generated content. While TikTok provides entertainment, creativity, and educational opportunities, it also exposes young users to potentially harmful and inappropriate material.

Traditional content moderation systems, designed to filter out unsuitable content, often rely on rule-based approaches that struggle with the growing influx of videos uploaded daily \cite{Grandinetti2023}. Furthermore, many existing methods, including those focusing on images \cite{taha2023armornet, taha2023inspectornet}, fail to account for the sequential nature of videos, which is crucial for effective content moderation. The diverse and inconsistent quality of user-generated videos, coupled with these limitations, exacerbates the challenge of automatic filtering \cite{taha2022filtering}. Despite ongoing efforts, many children still encounter content that is not appropriate for their age group, highlighting the inadequacy of current automated moderation methods \cite{weimann2023research, yang2022tik}.

Given the limitations of existing approaches, this study seeks to answer the following research question: \textit{How can we protect children from unsuitable content on TikTok?} To address this, we propose a novel solution leveraging state-of-the-art video classification models, aiming to improve the accuracy and robustness of content moderation systems in real-world scenarios.

The main contributions of this research are as follows:
\begin{enumerate}
    \item The potential of advanced transformer-based models for detecting unsuitable TikTok content for children is showcased.
    \item New techniques for robust content moderation are presented, improving the reliability of automated safety measures.
    \item Opportunities are opened for developers to create engaging and secure social media platforms tailored for children.
\end{enumerate}

In this paper, we present \textit{TikGuard}, a deep learning solution that utilizes advanced transformer-based architectures, demonstrating superior accuracy in detecting unsuitable TikTok content for children. By leveraging cutting-edge video classification models, this research sets a new benchmark for content moderation on social media platforms, ensuring a safer online experience for minors.

The rest of this paper is organized as follows: Section \ref{sec:relatedwork} discusses related work. Section \ref{sec:dataset} describes the dataset. Section \ref{sec:methodology} outlines the methodology. Section \ref{sec:results} presents the results, and Section \ref{sec:conclusion} concludes the paper.

\section{Related Works}
\label{sec:relatedwork}

This section reviews existing research on detecting inappropriate content in videos, particularly focusing on child safety. Various methodologies and datasets have been proposed to address this critical issue.

Shubham Singh et al. \cite{singh2019kidsguard} present "KidsGUARD," a fine-grained approach for detecting child unsafe content in videos, which addresses the challenge of sparsely located inappropriate frames in videos. To tackle this, the authors propose using an LSTM-based autoencoder to learn video representations from descriptors extracted by the VGG16 CNN. These encoded representations are then classified by an LSTM to detect child unsafe content. The methodology is evaluated on a substantial dataset of 109,835 video clips curated specifically for this task. The approach demonstrates the ability to detect inappropriate content with a granularity of one second, achieving an impressive recall of 81\% at a precision of 80\%, significantly outperforming traditional video encoding methods like Fisher Vector and VLAD.

Kanwal Yousaf et al. \cite{yousaf2022deep} present a deep learning-based approach to detect and classify inappropriate content in YouTube videos, focusing on child-oriented cartoon clips. The study introduces a manually annotated dataset of 111,156 cartoon video clips sourced from YouTube, categorized into safe, fantasy violence, and sexual-nudity classes. Using the EfficientNet-B7 model for feature extraction and a BiLSTM network for video representation, the proposed method achieves impressive accuracy (95.66\%) and an F1 score of 0.9267, outperforming traditional machine learning techniques and setting a new benchmark for inappropriate content detection in children's videos.

Dhiraj Murthy et al. \cite{murthy2024using} conducted a study to detect e-cigarette content in TikTok videos using computer vision techniques. They compiled a dataset of 826 still images from 254 TikTok posts, augmenting it with 89 images of white vapes and two support datasets containing over 9,000 images of random non-e-cigarette content. The researchers developed an object detection model based on YOLOv7, employing data augmentation techniques to improve the model's performance. The model achieved a recall of 0.77, precision of 0.863, and an F1 score of 0.814, demonstrating high accuracy in identifying vape devices, hands, and vapor, with significant reduction in false positives.

Several pre-trained models have been utilized for video classification tasks. Notably, \textbf{TimesFormer} \cite{bertasius2021space}, \textbf{VideoMAE} \cite{tong2022videomae}, and \textbf{ViViT} \cite{arnab2021vivit} have made significant strides in handling the complexities of video data. TimesFormer introduces factorized self-attention for managing temporal dependencies, VideoMAE employs masked autoencoders for efficient learning, and ViViT combines convolutional and transformer layers to capture both spatial and temporal features. These advancements have greatly enhanced video content moderation capabilities, especially on platforms like TikTok.

Our paper utilizes TimesFormer, VideoMAE, and ViViT to enhance the accuracy and efficiency of detecting inappropriate TikTok content for children.

\section{Dataset}
\label{sec:dataset}
The TikHarm dataset is a curated collection of TikTok videos specifically designed to train models for classifying harmful content. The dataset is formatted similarly to UCF101 \cite{soomro2012ucf101} but is tailored towards content accessible to children, with the objective of distinguishing between various types of potentially harmful material.

Data was meticulously gathered from TikTok, focusing on videos that are accessible to children to ensure that the dataset accurately reflects the type of content they are likely to encounter. The collected videos were manually labeled into four predefined categories:

\begin{itemize}
    \item \textbf{Harmful Content}: Videos that depict violence, dangerous actions that children might imitate, or other harmful behavior.
    \item \textbf{Adult Content}: Videos containing sexual content or other material deemed inappropriate for children.
    \item \textbf{Safe}: Videos that are appropriate and safe for children to view, such as popular cartoons.
    \item \textbf{Suicide}: Videos that depict, suggest, or discuss suicidal behavior or ideation.
\end{itemize}

The TikHarm dataset consists of 3,948 videos, divided into training, development (validation), and testing subsets. The duration and distribution statistics for each subset and class are detailed in Tables \ref{tab:subset_statistics} and \ref{tab:class_statistics}.

\begin{table}[ht]
\centering
\caption{Subset Statistics of the TikHarm Dataset}
\label{tab:subset_statistics}
\begin{tabular}{|c|c|c|c|}
\hline
\textbf{Subset} & \textbf{Samples} & \textbf{Avg Duration (s)} & \textbf{Total Duration (h)} \\
\hline
Train & 2762 & 38.71 & 29.71 \\
Dev & 790 & 38.57 & 4.24 \\
Test & 396 & 38.77 & 8.51 \\
\hline
\end{tabular}
\end{table}

\begin{table}[ht]
\centering
\caption{Class Distribution and Duration Statistics in the TikHarm Dataset}
\label{tab:class_statistics}
\begin{tabular}{|c|c|c|c|}
\hline
\textbf{Class} & \textbf{Samples} & \textbf{Avg Duration (s)} & \textbf{Total Duration (h)} \\
\hline
Safe & 997 & 65.36 & 18.1 \\
Adult & 977 & 36.25 & 9.84 \\
Harmful & 990 & 35.92 & 9.88 \\
Suicide & 984 & 16.96 & 4.63 \\
\hline
\end{tabular}
\end{table}

The annotation process was performed manually by a team of experts, ensuring high-quality labels that accurately reflect the content of each video. Annotators followed strict guidelines to categorize each video into one of the four predefined classes. This meticulous process ensures that the dataset is both reliable and effective for training robust video classification models.

\begin{figure}[h]
    \centering
    \begin{tabular}{cc}
        \includegraphics[height=2.5 cm]{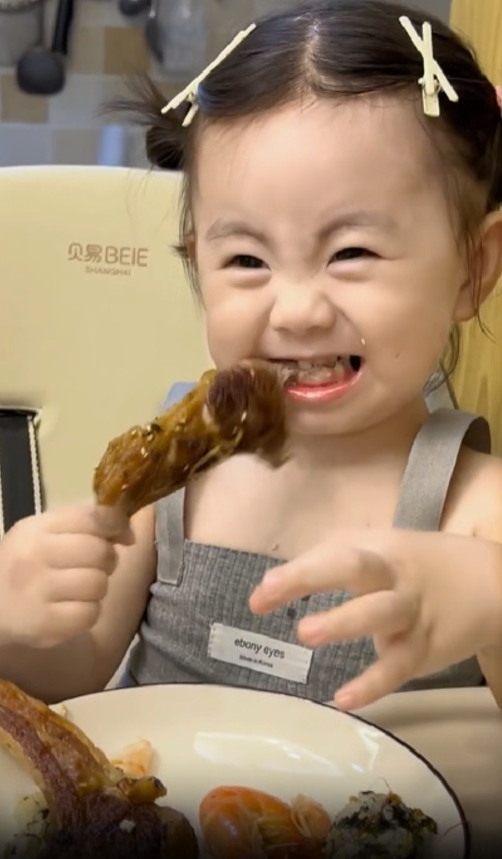} & \includegraphics[height=2.5cm]{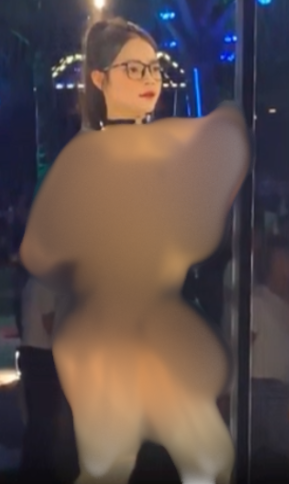} \\
        Safe & Adult Content \\
        \includegraphics[height=2.5 cm]{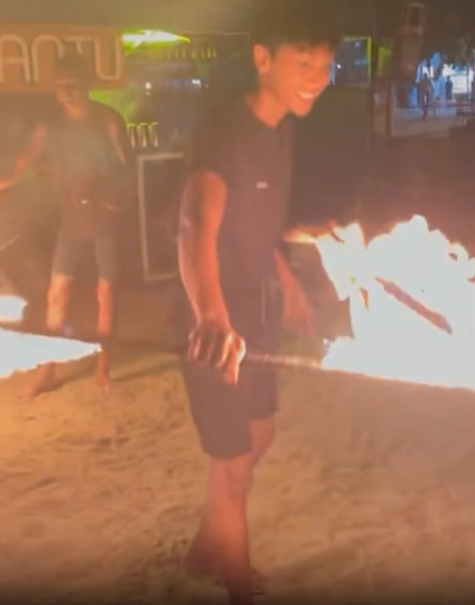} & \includegraphics[height=2.5 cm]{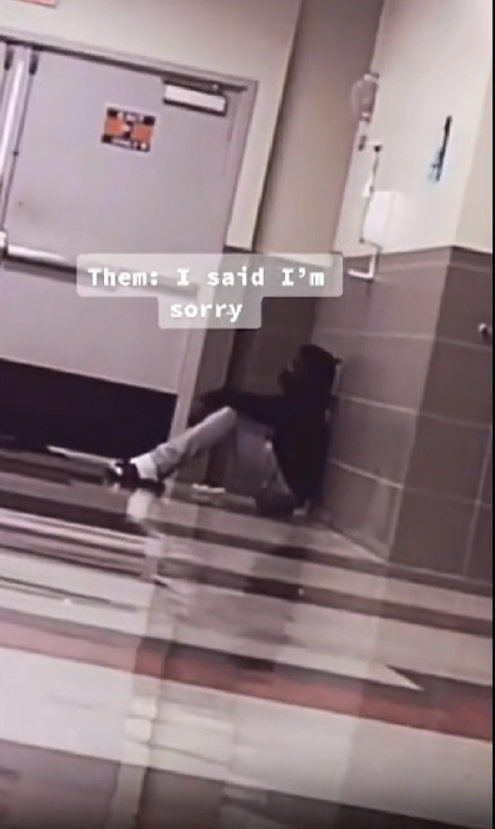} \\
        Harmful Content & Suicide \\
    \end{tabular}
    \caption{Examples of each class in the TikHarm dataset.}
    \label{fig:tikharm_classes}
\end{figure}

Figure \ref{fig:tikharm_classes} shows examples from each class in the TikHarm dataset.

The TikHarm dataset is invaluable for developing and evaluating video classification models aimed at automatically detecting and categorizing harmful content on social media platforms. Its focus on child-accessible content makes it a critical resource for enhancing the safety and moderation of digital content consumed by minors.

\section{Methodology}
\label{sec:methodology}
The proposed methodology leverages advanced transformer-based models to classify TikTok videos into predefined categories, ensuring the detection of unsuitable content for children.

\begin{figure}[ht]
    \centering
    \begin{tikzpicture}[
        node distance= .6cm and .6cm,
        box/.style={rectangle, draw, text width=2cm, align=center, minimum height=1cm, fill=blue!20},
        arrow/.style={-Stealth, thick, blue}
    ]

    \node[box] (node1) {Uniform Temporal Subsample};
    \node[box, right=of node1] (node2) {Normalization};
    \node[box, right=of node2] (node3) {Random Flips};
    \node[box, below=of node3] (node4) {Slide Scale};
    \node[box, left=of node4] (node5) {Resizing};
    \node[box, left=of node5] (node6) {Random Crop};

    \draw[arrow] (node1) -- node[above] {Data Flow} (node2);
    \draw[arrow] (node2) -- node[above] {Data Flow} (node3);
    \draw[arrow] (node3) -- node[right] {Data Flow} (node4);
    \draw[arrow] (node4) -- node[above] {Data Flow} (node5);
    \draw[arrow] (node5) -- node[above] {Data Flow} (node6);

    \end{tikzpicture}
    \caption{The preprocessing and augmentation pipeline}
    \label{fig:preprocessing_pipeline}
\end{figure}
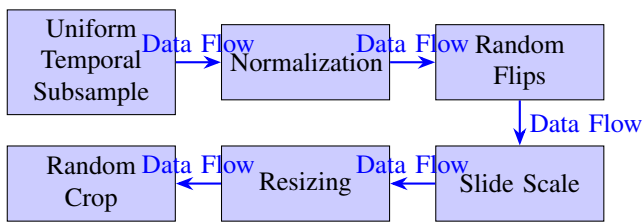

We designed a detailed preprocessing and augmentation pipeline to make the most of our video data, as shown in Figure \ref{fig:preprocessing_pipeline}. The first step was to extract frames from each video. Instead of using a fixed number of frames, we used a flexible method that adjusts based on the video's length and activity level. Videos with more action or fast-moving scenes had more frames extracted, while slower videos had fewer. This way, we made sure to capture the most important content by adapting the frame sampling rate to fit the video's pace.

Next, the frames were transformed by scaling pixel intensities to a float range of 0 to 1, followed by brightness normalization using specific mean and standard deviation values \cite{huang2023normalization}. We also applied geometric transformations, such as random horizontal flipping, which mirrors frames with a 50\% chance, and dynamic short-side scaling that maintains the aspect ratio by resizing the shorter edge to between 256 and 320 pixels. Finally, the frames were either resized or randomly cropped to match the target resolution, ensuring uniformity across different video sources. This thorough preprocessing process improves the generalizability and robustness of the data for deep learning tasks.

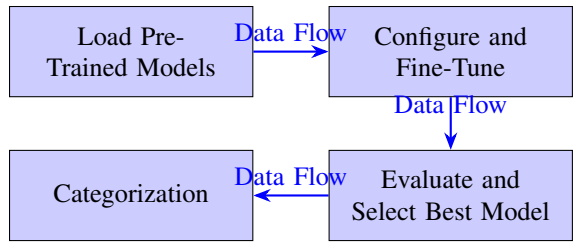
\begin{figure}[ht]
    \centering
    \begin{tikzpicture}[
        node distance= .7cm and 1cm,
        box/.style={rectangle, draw, text width=3cm, align=center, minimum height=1.2cm, fill=blue!20},
        arrow/.style={-Stealth, thick, blue}
    ]

    \node[box] (node1) {Load Pre-Trained Models};
    \node[box, right=of node1] (node2) {Configure and Fine-Tune};
    \node[box, below=of node2] (node3) {Evaluate and Select Best Model};
    \node[box, left=of node3] (node4) {Categorization};

    \draw[arrow] (node1) -- node[above] {Data Flow} (node2);
    \draw[arrow] (node2) -- node[above] {Data Flow} (node3);
    \draw[arrow] (node3) -- node[above] {Data Flow} (node4);

    \end{tikzpicture}
    \caption{The model fine-tuning process}
    \label{fig:model_finetuning_pipeline}
\end{figure}

We fine-tuned three state-of-the-art transformer-based models—Timesformer, VIVIT, and VideoMAE—using pre-trained weights to classify TikTok videos and detect inappropriate content for children. These models were adjusted to our specific objectives and optimized for performance. The overall fine-tuning process is shown in Figure \ref{fig:model_finetuning_pipeline}.

Key hyperparameters such as learning rates, batch sizes, and epochs were optimized for our dataset. Specifically, we set the learning rate to 5e-5, the batch size to 4, and the number of epochs to 9. We also used a warmup ratio of 0.1 and capped the maximum training steps at 6905. To improve efficiency, we employed mixed precision training and regularly performed validation to avoid overfitting \cite{micikevicius2017mixed}.

The best model, selected based on validation accuracy, was then tested on a separate dataset to ensure it could generalize well. This extensive evaluation confirmed the model’s ability to detect inappropriate content reliably, contributing to a safer online environment for children.

After fine-tuning, we evaluated the models using accuracy, precision, recall, and F1 score metrics. \textbf{Accuracy} measures how often the model correctly classifies instances overall:

\begin{equation}
\text{Accuracy} = \frac{TP + TN}{TP + TN + FP + FN}
\label{eq:accuracy}
\end{equation}

\textbf{Precision} is the proportion of true positives among all predicted positives:

\begin{equation}
\text{Precision} = \frac{TP}{TP + FP}
\label{eq:precision}
\end{equation}

\textbf{Recall} (or Sensitivity) measures how well the model identifies true positives from all actual positives:

\begin{equation}
\text{Recall} = \frac{TP}{TP + FN}
\label{eq:recall}
\end{equation}

\textbf{F1 Score} is the harmonic mean of precision and recall, providing a balanced metric:

\begin{equation}
\text{F1 Score} = 2 \times \frac{\text{Precision} \times \text{Recall}}{\text{Precision} + \text{Recall}}
\label{eq:f1score}
\end{equation}

These metrics were chosen to provide a comprehensive evaluation of the model’s performance in detecting unsuitable content.

\section{Results}
\label{sec:results}
In this section, we present the performance results of our proposed Transformer-based models on the TikHarm dataset, focusing on the detection of unsuitable TikTok content for children.

\begin{table}[ht]
\centering
\caption{Performance of models on the Validation Set}
\label{tab:validation}
\begin{tabular}{|l|c|c|c|c|}
\hline
\textbf{Model} & \textbf{ACC} & \textbf{F1} & \textbf{Recall} & \textbf{Precision} \\
\hline
\textbf{TimesFormers} & \textbf{0.8666} & \textbf{0.8662} & \textbf{0.8679} & \textbf{0.8662} \\
\textbf{VideoMAE}     & 0.7911 & 0.7917 & 0.7915 & 0.7911 \\
\textbf{VIVIT}        & 0.8616 & 0.8624 & 0.8646 & 0.8624 \\
\hline
\end{tabular}
\end{table}

\begin{table}[ht]
\centering
\caption{Performance of models on the Test Set}
\label{tab:test}
\begin{tabular}{|l|c|c|c|c|}
\hline
\textbf{Model} & \textbf{ACC} & \textbf{F1} & \textbf{Recall} & \textbf{Precision} \\
\hline
\textbf{TimesFormers} & \textbf{0.8671} & \textbf{0.8668} & \textbf{0.8671} & \textbf{0.8669} \\
\textbf{VideoMAE}     & 0.7816 & 0.7802 & 0.7816 & 0.7826 \\
\textbf{VIVIT}        & 0.8418 & 0.8408 & 0.8418 & 0.8467 \\
\hline
\end{tabular}
\end{table}

The results indicate that TimesFormers consistently outperforms the other models across both the validation and test sets in terms of accuracy, F1 score, recall, and precision. Specifically, TimesFormers achieved an accuracy of 0.8666 on the validation set (Table~\ref{tab:validation}) and 0.8671 on the test set (Table~\ref{tab:test}), demonstrating its robustness and reliability in detecting unsuitable TikTok content for children.

VideoMAE, on the other hand, showed the lowest performance among the three models, with an accuracy of 0.7911 on the validation set (Table~\ref{tab:validation}) and 0.7816 on the test set (Table~\ref{tab:test}). Although its F1 score and precision are relatively close to its accuracy, VideoMAE's performance suggests that it may not be as effective in capturing the nuances of harmful content as the other models.

VIVIT performed well, achieving an accuracy of 0.8616 on the validation set (Table~\ref{tab:validation}) and 0.8418 on the test set (Table~\ref{tab:test}). While it did not surpass TimesFormers, VIVIT's results indicate that it is a competitive model capable of effectively identifying unsuitable content.

The higher performance metrics of TimesFormers can be attributed to its advanced temporal modeling capabilities, which are crucial for understanding the context in video sequences. VIVIT also leverages temporal information effectively, but it appears that TimesFormers has a slight edge in this aspect. VideoMAE's lower performance may be due to its architectural differences and possibly less effective handling of temporal dependencies in the video data.

\section{Conclusion}
\label{sec:conclusion}

In conclusion, this paper introduced TikGuard, a transformer-based approach utilizing TimesFormers, VideoMAE, and ViVit models for detecting unsuitable TikTok content for children, thereby addressing the pressing question: How can we safeguard young users from harmful online content? By harnessing the power of the Tikharm dataset, we demonstrated the superiority of TimesFormers with an accuracy of 86.7\%, showcasing the potential of transformer-based architectures in tackling the challenge of detecting harmful content on social media platforms like TikTok. However, the quest for a safer online environment for children is far from over. To further enhance TikGuard's performance, future work should focus on improving model robustness, exploring additional transformer architectures, expanding the dataset to cover a broader range of unsuitable content categories, and incorporating multimodal information, such as audio and text, for a more comprehensive understanding of video content. By doing so, we can continue to pave the way for a more responsible and child-friendly social media landscape, ultimately ensuring that the digital world remains a safe and enriching space for our children to learn, grow, and explore.

\bibliographystyle{IEEEtran}
\bibliography{main}
\end{document}